\documentclass{article}
\usepackage{spconf,amsmath,graphicx}

\usepackage{times}
\usepackage{epsfig}
\usepackage{graphicx}
\usepackage{amsmath}
\usepackage{bbm, dsfont}
\usepackage{amssymb}
\usepackage{booktabs}
\usepackage[dvipsnames, table]{xcolor}
\definecolor{Gray}{gray}{0.85}
\usepackage{multirow}
\usepackage{pifont}
\usepackage{algorithm}
\usepackage{algpseudocode}
\usepackage{comment}
\usepackage{hyperref}
\usepackage{subcaption}
\usepackage{caption}
\usepackage{tikz}
\usepackage{comment}
\usepackage{amsmath,amssymb} 
\usepackage{color}
\usepackage{amsthm}
\newtheorem{theorem}{Theorem}

\title{Designing strong baselines for ternary neural network quantization through support and mass equalization}
\name{Edouard Yvinec$^{1,2}$, Arnaud Dapogny$^2$, Kevin Bailly$^{1,2}$ \thanks{This work has been supported by the french National Association for Research and Technology (ANRT), the company Datakalab (CIFRE convention C20/1396) and by the French National Agency (ANR)  (FacIL, project ANR-17-CE33-0002). This work was granted access to the HPC resources of IDRIS under the allocation 2022-AD011013384 made by GENCI.}}
\address{Sorbonne Université$^1$, CNRS, ISIR, f-75005, 4 Place Jussieu 75005 Paris, France \and Datakalab$^2$, 114 boulevard Malesherbes, 75017 Paris, France}
\begin{document}
%
\maketitle
\begin{abstract}
Deep neural networks (DNNs) offer the highest performance in a wide range of applications in computer vision. These results rely on over-parameterized backbones, which are expensive to run. This computational burden can be dramatically reduced by quantizing (in either data-free (DFQ), post-training (PTQ) or quantization-aware training (QAT) scenarios) floating point values to ternary values (2 bits, with each weight taking value in $\{-1,0,1\}$). In this context, we observe that rounding to nearest minimizes the expected error given a uniform distribution and thus does not account for the skewness and kurtosis of the weight distribution, which strongly affects ternary quantization performance.
This raises the following question: shall one minimize the highest or average quantization error? To answer this, we design two operators: TQuant and MQuant that correspond to these respective minimization tasks. We show experimentally that our approach allows to significantly improve the performance of ternary quantization through a variety of scenarios in DFQ, PTQ and QAT and give strong insights to pave the way for future research in deep neural network quantization.
\end{abstract}
\section{Introduction}
As the performance of deep neural network grows, so do their computational requirements: in computer vision, popular architectures such as ResNet \cite{he2016deep}, MobileNet V2 \cite{sandler2018mobilenetV2} and EfficientNet \cite{tan2019efficientnet} rely on high expressivity from millions of parameters to effectively tackle challenging tasks such as classification \cite{imagenet_cvpr09}, object detection \cite{pascal-voc-2012} and segmentation \cite{cordts2016cityscapes}. 

In order to deploy these models using lower power consumption and less expensive hardware, many compression techniques have been developed to reduce the latency and memory footprint of large convolutional neural networks (CNN). Quantization is one of the most efficient of these techniques and consists in converting floating values with large bit-width to fixed point values encoded on lower bit-width. Quantization can be performed in three major contexts. First, data-free quantization (DFQ) \cite{nagel2019data,meller2019same,squant2022,yvinec2022rex,yvinec2023powerquant}, where, quantization is applied without data, hopefully not degrading the model accuracy. Second, post training quantization (PTQ) \cite{banner2019post,nagel2020up,li2021brecq}: in this setup, one seeks to tune the quantization operator given an already trained model and a calibration set (usually a fraction of the training set). Third, quantization aware training (QAT) \cite{wu2018training,jacob2018quantization,achterhold2018variational,louizos2018relaxed,sheng2018quantization,ullrich2017soft,zhou2016dorefa}, in which, given a model and a training set, custom gradient descent proxies are usually implemented so as to circumvent zero gradients that stems from the rounding operation.

\begin{figure*}[!t]
    \centering
    \includegraphics[width = \linewidth]{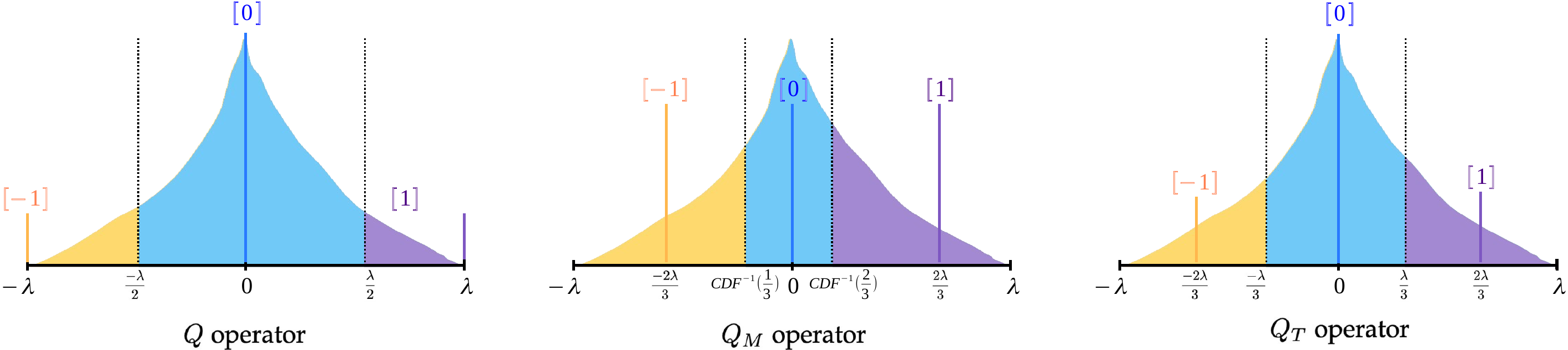}
     \caption{Comparison of the ternary distribution (colored Dirac distributions) from different quantization operators: the naive quantization operator $Q$, the mass balancing operator $Q_M$ and the proposed ternary operator $Q_T$.}
    \label{fig:compare_quantif}
\end{figure*}

While noawadays most methods are successful in quantizing models to 8 or 4 bits while preserving the accuracy \cite{park2018value,nagel2020up,yvinec2023spiq,yvinec2023powerquant}, lower bit-widths remain an open challenge. In particular, ternary quantization, where weights are mapped to $\{-1,0,1\}$ \cite{zhu2016trained,banner2019post} is an extreme form of neural network quantization, that comes with its specific challenges. While most quantization techniques rely on a rounding operation \cite{park2018value,nagel2019data,meller2019same,yvinec2022rex}, they struggle to preserve high accuracies when applied in the ternary quantization regime. In this work, we show that this limitation is at least partly due to the distribution of the weights: in fact, assuming a bell-shaped distribution \cite{nagel2019data,yvinec2022red++}, the tails of the distribution, which represent very few parameters, will be quantized to non-zero values. Consequently, the naive rounding operation erases most of the learned operations which leads to huge accuracy drops.
Based on this observation, we conclude that the naive quantization operator, that amounts to minimizing the expected error given a uniform distribution (round to nearest) is particularly ill-suited in ternary quantization. This leads to the following question: \textit{shall one minimize the highest quantization error or the average quantization error on the weights?} To answer this question we design two quantization operators: TQuant, which minimizes the highest quantization error, and MQuant, which minimizes the average quantization error assuming a bell shaped distribution. In Figure \ref{fig:compare_quantif}, we illustrate these two operators. We empirically show that these operators achieve strong improvements in a broad range of scenarios for DFQ, PTQ and QAT.

\section{Methodology}
Let's consider $F$ a network with $L$ layers and weights $W_l$ for each layer indexed by $l$. We note $Q$ the $b$-bits quantization operator that quantizes the weights $W_l$ from $[\min\{W_l\}; \max\{W_l\}] \subset \mathbb{R}$ to the quantization interval $ [- 2^{b-1} ; 2^{b-1} -1] \cap \mathbb{Z}$ and $W_l^q$ the quantized weights and is defined as:
\begin{equation}\label{eq:quantization_operator}
    (W^q) = Q(W) = \left\lfloor\frac{W}{\lambda}\right\rceil
\end{equation}
where $\lfloor\cdot\rceil$ denotes the rounding operation and $\lambda$ is a row-wise rescaling factor selected such that each coordinate of $W^q$ falls in the quantization interval, \textit{i.e.} $(\lambda)_i$ is the maximum between $\frac{|\min\{(W)_i\}|}{2^{b-1}}$ and $\frac{|\max\{(W)_i\}|}{2^{b-1}-1}$.
When the number of bits $b$ is large, e.g. $b=4$ or $b=8$, most of the scalar values $w$ of $W$ will not be quantized to the extreme values of the quantized interval $[- 2^{b-1} ; 2^{b-1} -1]$ and as such not considering the tail of the distribution does not affect performance. This is not true when $b=2$ (ternary) as can be seen in Figure \ref{fig:compare_quantif} (a). Only the tails of the distribution are quantized to non-zero values which represent very few values. To reduce this effect, we propose to quantization operators, namely TQuant ($Q_T$) and MQuant ($Q_M$), that either leads to minimize the highest scalar quantization error or to equalize the support of each ternary value, respectively.

\subsection{Support Equalization}
We define the ternary quantization operator $Q_T$ such that we have an equality of the measure of each pre-image of $-1, 0$ and $1$, \textit{i.e.}
\begin{equation}\label{eq:partition_Q_ternary}
    \begin{cases}
        Q_T^{-1}(\{-1\}) = \left[-\lambda;-\frac{\lambda}{3}\right]\\
        Q_T^{-1}(\{0\}) = \left[-\frac{\lambda}{3};\frac{\lambda}{3}\right]\\
        Q_T^{-1}(\{1\}) = \left[\frac{\lambda}{3};\lambda\right]\\
    \end{cases}
\end{equation}
Consequently, all quantized values are mapped from ranges equal in size, a third of the original support. This is achieved by using two thirds of the scaling factor instead, in equation \ref{eq:quantization_operator}. Then, the maximal error between any scalar $w\in W$ and the quantization $Q^{-1}(Q(w))$ is $\frac{\lambda}{3}$. 
This operator balances the support of the original weight values with the quantization space $\{-1,0,1\}$. Thus $Q_T$ minimises the maximal error per weight value. However it doesn't balance the mass. More formally the number of weight values assigned to $-1$ is not necessarily equal to the number of values assigned to $1$ nor $0$. To achieve the mass equalization, we introduce MQuant.

\subsection{Mass Equalization}
For each value $v\in \{-1,0,1\}$ the spaces $Q_M^{-1}(\{v\})$ have equal mass. This is achieved using $\lambda \times \frac{5}{7\sqrt{2}}$ as a scaling factor.
\begin{theorem}\label{expectederror}
Under a centered Gaussian prior, the $Q_M$ operator minimizes the expected error per scalar weight value.
\end{theorem}
We provide the proof in Appendix \ref{sec:appendix_proof_thm4}. This theoretical result allows us to compare the behaviour of the error in ternary quantization. \textit{Which is more important, minimizing the maximum error or the expected error from quantization?} In what follows, we provide empirical answers to this question, and show that both operators lead to significantly more robust baselines for ternary quantization.

\section{Experiments}
We evaluate the proposed quantization operators in three contexts: quantization without data (data-free), quantization with a calibration set (PTQ) and quantization with full dataset (QAT). To provide insightful results for real world applications, we consider standard CNN backbones on ImageNet \cite{imagenet_cvpr09}, such as ResNet 50 \cite{he2016deep}, MobileNet V2 \cite{sandler2018mobilenetV2} and EfficientNet B0 \cite{tan2019efficientnet}. In the context of tiny machine learning, we also considered ResNet 8 on Cifar10.
\subsection{Data-Free Quantization}
For our evaluation in the context of data-free quantization, we compare our two operators TQuant and MQuant with the baseline SQuant \cite{squant2022} that achieves state-of-the-art accuracy in data-free quantization. In a similar vein as in \cite{yvinec2022rex} we evaluate our approach using different expansions orders that allow to find several accuracy-computational burden trade-offs.

\begin{figure}[!t]
    \centering
    \includegraphics[width = \linewidth]{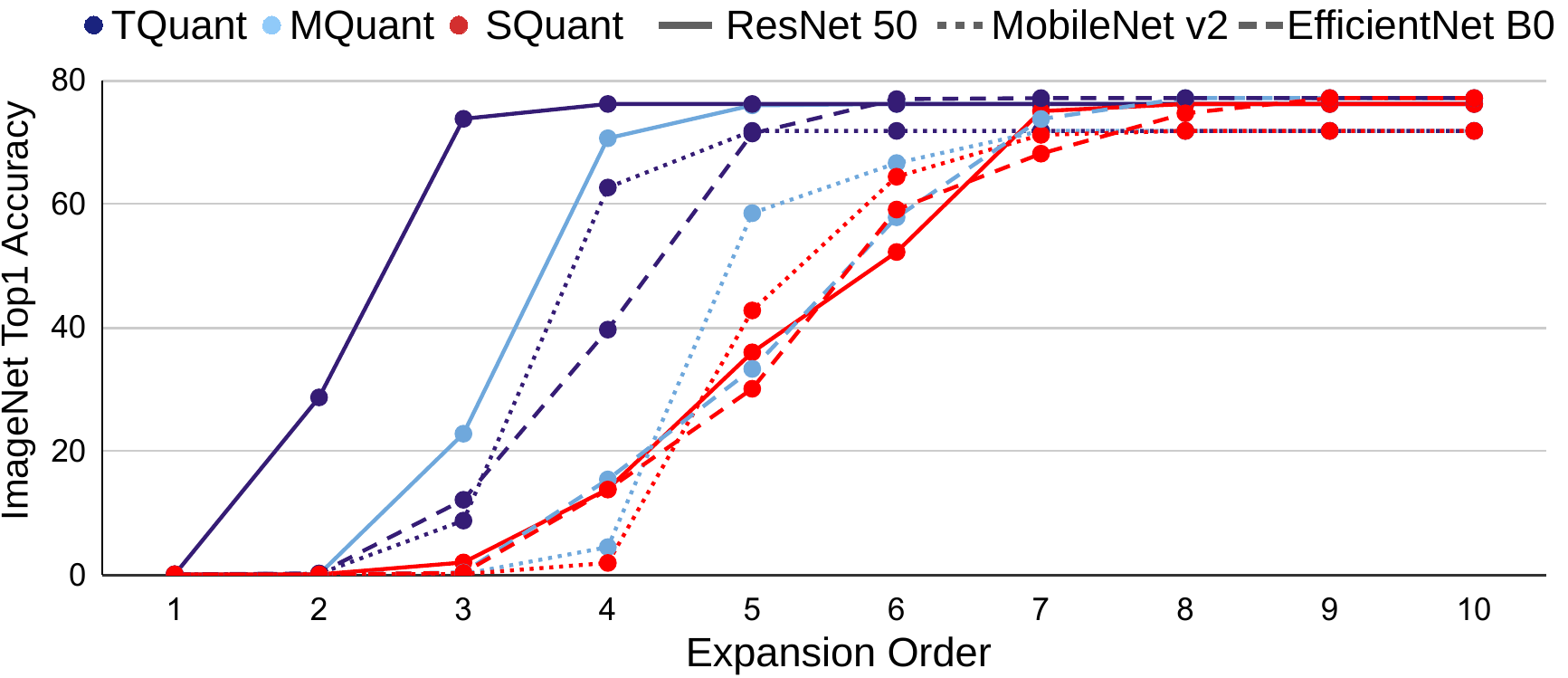}
    \caption{Comparison between the proposed TQuant, MQuant and the state-of-the-art SQuant \cite{squant2022} operator for ternary weights expansion and fixed 8 bits activations (W2/A8). The ImageNet top1 accuracy is reported for ResNet 50, MobileNet V2 and EfficientNet B0.}
    \label{fig:q_operators_exp}
\end{figure}
In Figure \ref{fig:q_operators_exp}, we compare the influence of the quantization operator on the accuracy for ternary weight quantization. We observe across all three tested architectures that both MQuant and TQuant offer higher accuracies than SQuant by a significant margin. For instance, an expansion of order $3$ with TQuant on ResNet 50 offers $71.72$ points higher accuracy and MQuant offers $58.83$ higher accuracy for an order 4 expansion. In parallel, TQuant reaches full-precision accuracy with half the expansion of SQuant and $50\%$ less than MQuant which shows the better performance of the proposed method. This suggests that the proposed ternary-specific operators TQuand and MQuant enable overall stronger baselines than SQuant.

\begin{figure}[!t]
    \centering
    \includegraphics[width = \linewidth]{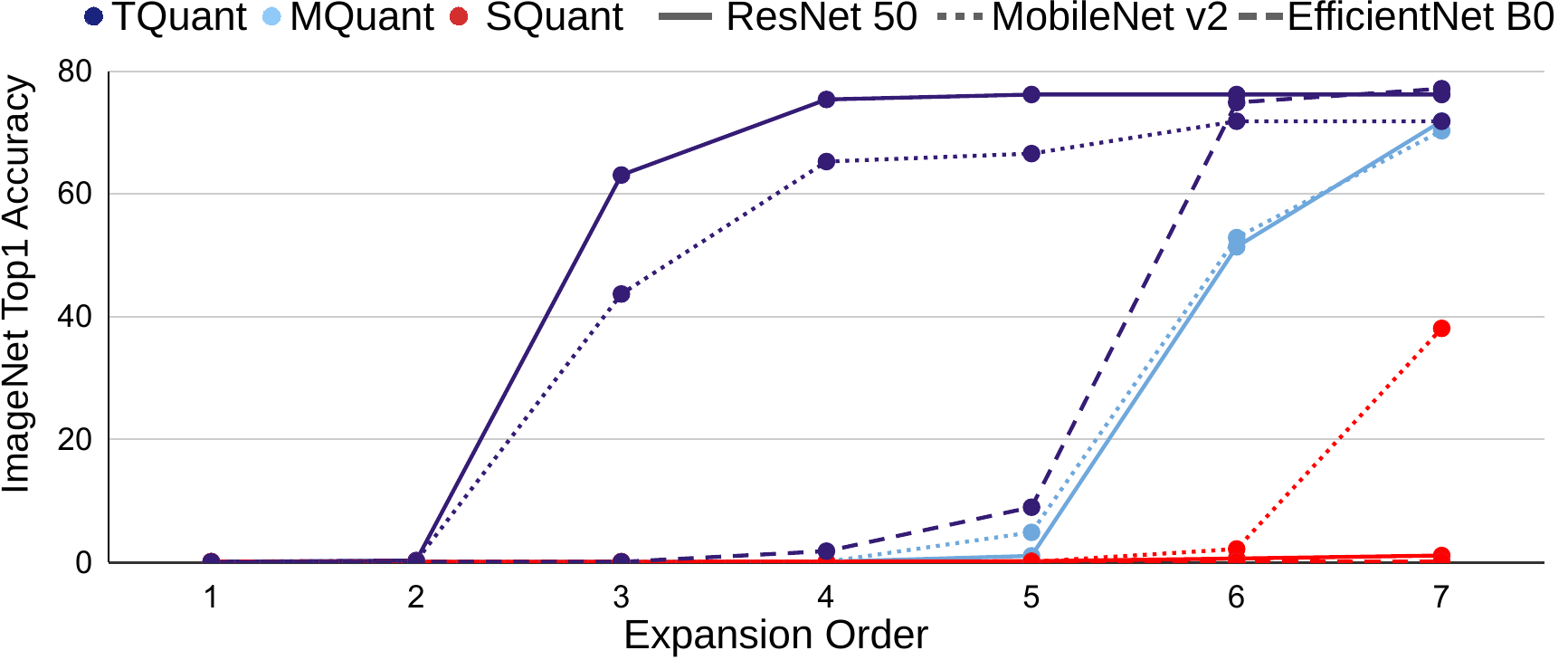}
    \caption{Comparison between the proposed TQuant, MQuant and the state-of-the-art SQuant \cite{squant2022} operator for ternary weights and activation expansion (W2/A2).}
    \label{fig:inputs_exp}
\end{figure}
In Figure \ref{fig:inputs_exp}, we apply the expansion quantization to both weights and activations. We observe very similar results. For instance, on MobileNet V2 and EfficientNet B0, only TQuant reaches the full-precision accuracy in a reasonable expansion order. It is worth noting that MQuant is significantly less efficient on activation quantization due to the lack of data. Formally, the scaling factors are derived from the standard deviations stored in the batch normalization layers which give a much better estimation of the maximum values (TQuant) than the cdf (MQuant). In a data-driven context, MQuant would be very similar to the method applied in \cite{li2016ternary}. However, TQuant doesn't suffer from this limitation and vastly outperforms SQuant as a quantization operator for data-free quantization in ternary representation.

\subsection{Post-Training Quantization}
We modified the two most popular methods: AdaRound \cite{nagel2020up} and BrecQ \cite{li2021brecq} to use with both MQuant and TQuant for weight ternary quantization (W2/A32). These techniques consist in learning, for each scalar value, whether it shall be rounded up or down. The procedure is agnostic to the scaling factor $\lambda$. In our results, the rounding operation is determined by the PTQ method while the scales are given by the operator.
In Table \ref{tab:ptq_results}, we report our results on PTQ. First, we observe marginal slow down in terms of processing time when using our custom operators which is due to the added multiplicative term on the weight scaling. It is worth noting that the resulting quantized models will all share the exact same runtime.
Second, we observe significant accuracy improvement when using TQuant on both PTQ methods, adding $28.7$\% and $32.22$\% accuracy with AdaRound and BrecQ respectively. Furthermore, it appears that MQuant even outperforms TQuant which suggests that when using a calibration sets, minimizing the average quantization error offers better performance than minimizing the highest quantization error.

\begin{table}[!t]
\caption{Accuracy results of a quantized (W2/A32) ResNet 8 for Cifar10. The process is performed on a RTX 2070. We report the quantization pre-processing time in minutes (m).}
\label{tab:ptq_results}
\centering
\setlength\tabcolsep{4pt}
\begin{tabular}{|c|c|c|c|}
\hline
PTQ method & operator & accuracy & Processing Time\\
\hline
 -& - & 89.100 & -\\
\hline
\multirow{3}{*}{AdaRound} & native & 11.790 $\pm$ 3.210 & 5m01 \\
 & MQuant & \textbf{42.910} $\pm$ 0.620 & 5m18 \\
 & TQuant & 40.490 $\pm$ 0.250 & 5m18 \\
\hline
\multirow{3}{*}{BrecQ} & native & 25.780 $\pm$ 2.440 & 3m45 \\
 & MQuant & \textbf{63.540} $\pm$ 0.850 & 3m50 \\
 & TQuant & 58.000 $\pm$ 1.120 & 3m50 \\
\hline
\end{tabular}
\end{table}

\subsection{Quantization Aware Training}
\begin{table}[!t]
\caption{Accuracy results of ResNet 8 for Cifar10, quantized in W2/A4 using straight through estimation \cite{yinunderstanding}.}
\label{tab:qat_results}
\centering
\setlength\tabcolsep{4pt}
\begin{tabular}{|c|c|c|c|}
\hline
 & Baseline & MQuant & TQuant \\
\hline
accuracy & 42.910  $\pm$ 14.61  & 68.250 $\pm$ 6.26 & \textbf{82.620} $\pm$ 2.43 \\
\hline
\end{tabular}
\end{table}

In QAT, the quantized weights and their scales are learned from scratch. In practice, the weights are kept in full-precision (float32) during the training process. The forward pass uses their quantized version while the backward updates the full-precision values. In our tests, the quantized values applied in the forward passes are derived using the naive rounding operator (eq \ref{eq:quantization_operator}) and scales from the given operator.
In Table \ref{tab:qat_results}, we report our results for ResNet 8 on Cifar10. We observe that both TQuant and MQuant improve the accuracy of the resulting model with TQuant almost enabling full-precision accuracy. TQuant also stabilizes the result as the variance on the accuracy drops from the original $14.610$\% down to $2.429$\%. We conclude that, in QAT, minimizing the highest quantization error provides the best results for ternary quantization.

From all our experiments in different quantization context, we deduce several insights on ternary quantization:
\begin{itemize}
    \item when \textit{no data} is available, we should minimize the \textit{maximum} quantization error (TQuant),
    \item when \textit{few data} are available, we should minimize the \textit{average} quantization error (MQuant),
    \item when \textit{all the data} is available, we should minimize the \textit{maximum} quantization error (TQuant).
\end{itemize}

\section{Conclusion}
In this work, we investigated ternary quantization of deep neural networks. This very low bit representation offers massive inference speed-ups and energy consumption reduction enabling deployment on edge devices of large neural networks but at the cost of the accuracy. We highlight that part of the accuracy degradation arises from the rounding operation itself which is unsuitable for preserving the predictive function. To tackle this limitation, we propose two quantization operator, namely TQuant and MQuant, that respectively minimize the highest and average quantization errors assuming the correct empirical prior. Extensive testing in data-free quantization, post-training quantization and quantization-aware training contexts show that the proposed ideas, albeit simple, leads to designing stronger baselines for ternary deep neural network compression as compared with existing state-of-the-art methods.

\appendix
\section{Proof of Theorem 1}\label{sec:appendix_proof_thm4}
\begin{proof}
Let $w$ be a scalar value from a symmetric distribution $W$ with support $]-\lambda; \lambda[$. A quantization operator $Q^q_a$ is defined by two parameters $a$ and $q$ such that
\begin{equation}
    Q^q_a : w\mapsto \begin{cases}
        q & \text{ if } w \geq a\\
        0 & \text{ if } w \in ]-a;a[\\
        -q & \text{ otherwise}\\
    \end{cases}
\end{equation}
We want to solve the following minimization problem:
\begin{equation}
    \min_{a,q} \mathbb{E}\left[ \| Q^q_a(w) - w\| \right]
\end{equation}
We develop the expression $\mathbb{E}\left[ \| Q^q_a(w) - w\| \right]$ in a sum of integrals
\begin{equation}
\begin{aligned}
    \mathbb{E}\left[ \| Q^q_a(w) - w\| \right] =& \int_{-\lambda}^{-a} \|w+q\|d\mathbb{P}(w) \\
    &+ \int_{-a}^{a} \|w\|d\mathbb{P}(w) + \int_{a}^{\lambda} \|w-q\|d\mathbb{P}(w)
\end{aligned}
\end{equation}
Assuming, $\|\dot\|$ is the quadratic error, then $ \int_{a}^{\lambda} \|w-q\|d\mathbb{P}(w)$ is minimized by $q=\mathbb{E}[\mathbbm{1}_{[-\lambda;-a]}w]$.
By hypothesis, the distribution is symmetrical, \textit{i.e.} $\mathbb{P}(w)$ is even. Therefore $\mathbb{E}[\mathbbm{1}_{[-\lambda;-a]}w] = - \mathbb{E}[\mathbbm{1}_{[a;\lambda]}w]$ and 
\begin{equation}
    \mathbb{E}\left[ \| Q^q_a(w) - w\| \right] = \mathbb{V}[\mathbbm{1}_{[-a;a]}w] + 2\mathbb{V}[\mathbbm{1}_{[a;\lambda]}w]
\end{equation}
Assuming a Gaussian prior over $W$ (centered and reduced for the sake of simplicity), the variance terms correspond to variances of truncated Gaussian distribution \cite{johnson1995continuous}:
\begin{equation}
    \begin{cases}
        \mathbb{V}[\mathbbm{1}_{[-a;a]}w]= 1 - \frac{2 a \phi(a)}{2\Phi(a)-1}\\
        \mathbb{V}[\mathbbm{1}_{[a;\lambda]}w]= 1 - \frac{\lambda \phi(\lambda) - a \phi(a)}{\Phi(\lambda)-\Phi(a)} - \left(\frac{\phi(\lambda) - \phi(a)}{\Phi(\lambda)-\Phi(a)}\right)^2
    \end{cases}
\end{equation}
where $\phi$ is the density function of the Gaussian distribution and $\Phi$ the cumulative distribution function. To solve the minimization problem, we search for the critical points in $a$.
We recall that $\phi'(a) = -2a\phi(a)$ and $\Phi'(a) = \phi(a)$. 
\begin{equation}\resizebox{\hsize}{!}{%
    $
\begin{aligned}
    \frac{\partial \mathbb{E}\left[ \| Q^q_a(w) - w\| \right]}{\partial a} &= 2\phi(a)\frac{(2a^2+1)(2\Phi(a) - 1) - 2a\phi(a)}{\left(2\Phi(a)-1\right)^2}\\
    &+ \phi(a)\frac{\lambda \phi(\lambda) - a \phi(a) - (2a^2+1)(\Phi(\lambda)-\Phi(a))}{\left(\Phi(\lambda)-\Phi(a)\right)^2}\\
    &+ \phi(a)\frac{\phi(\lambda)-\phi(a) - 2a (\Phi(\lambda)-\Phi(a))}{\left(\Phi(\lambda)-\Phi(a)\right)^2}\frac{\phi(\lambda) - \phi(a)}{\Phi(\lambda)-\Phi(a)}
\end{aligned}
$}
\end{equation}
To solve $\frac{\partial \mathbb{E}\left[ \| Q^q_a(w) - w\| \right]}{\partial a} = 0$ for $a$, we simplify the equation to get
\begin{equation}\resizebox{\hsize}{!}{%
    $
    \begin{aligned}
    2\frac{(2a^2+1)(2\Phi(a) - 1) - 2a\phi(a)}{\left(2\Phi(a)-1\right)^2}\left(\Phi(\lambda)-\Phi(a)\right)^2-\frac{(\phi(\lambda) - \phi(a))^2}{\Phi(\lambda)-\Phi(a)}\\
    = (\lambda-2a) \phi(\lambda) + a \phi(a) - (2a^2+1)(\Phi(\lambda)-\Phi(a))
\end{aligned}
$}
\end{equation}
To solve this equation, we assume that $\Phi(\lambda) - \Phi(a) \approx \Phi(-a)$ and $\phi(\lambda)\approx 0$. Consequently, we obtain the second order polynomial: $2a^2+1 - 5a\phi(a) - 3\phi(a)^2=0$. We deduce, by solving the polynomial in $a$ and in $\phi(a)$:
\begin{equation}
    \begin{cases}
        a = \frac{1}{4} \left(\sqrt{49 \phi(a)^2 - 8} + 5\phi(a)\right)\\
        \phi(a) = \frac{1}{6}\left(\sqrt{12+49a^2} -5a \right) \\
    \end{cases}
\end{equation}
We find $a=\frac{5}{7\sqrt{2}}$ and as such $\Phi^{-1}(a) \approx \frac{2}{3}$ which corresponds to the definition of $Q_M$.
\end{proof}

\bibliographystyle{IEEEbib}
\bibliography{output}

\end{document}